\documentclass{article}
\usepackage{ijcai20}
\usepackage{times}
\usepackage{soul}
\usepackage{url}
\usepackage[hidelinks]{hyperref}
\usepackage[utf8]{inputenc}
\usepackage[small]{caption}
\usepackage{graphicx}
\usepackage{amsmath}
\usepackage{amsthm}
\usepackage{booktabs}
\usepackage{algorithm}
\usepackage{algorithmic}
\usepackage{booktabs}
\usepackage{amssymb}
\usepackage{multirow}
\usepackage{pifont}
\usepackage{subfigure}   

\title{Self-Supervised Gait Encoding with Locality-Aware Attention \\ for Person Re-Identification }

\author{
Haocong Rao$^{1,3,5}$\footnotemark[1]
\and
Siqi Wang$^2$\footnotemark[1]\and
Xiping Hu$^{1,4,5}$\and
Mingkui Tan$^3$\and
Huang Da$^2$\and \\
Jun Cheng$^{1,5}$\And
Bin Hu$^4$
\affiliations
$^1$Shenzhen Institutes of Advanced Technology, Chinese Academy of Sciences\\
$^2$National University of Defense Technology\\
$^3$South China University of Technology\\
$^4$Lanzhou University\\
$^5$The Chinese University of Hong Kong, Hong Kong
\emails
haocongrao@gmail.com,
wangsiqi10c@nudt.edu.cn,
\{xp.hu, jun.cheng\}@siat.ac.cn
mingkuitan@scut.edu.cn,
huangda1109@163.com,
bh@lzu.edu.cn
}

\begin{document}
\maketitle
\footnotetext[1]{Authors contribute equally.}
\begin{abstract}

Gait-based person re-identification (Re-ID) is valuable for safety-critical applications, and using only \textit{3D skeleton data} to extract discriminative gait features for person Re-ID is an emerging open topic. Existing methods either adopt hand-crafted features or learn gait features by traditional supervised learning paradigms. Unlike previous methods, we for the first time propose a generic gait encoding approach that can utilize \textit{unlabeled} skeleton data to learn gait representations in a \textit{self-supervised} manner. Specifically, we first propose to introduce self-supervision by learning to reconstruct input skeleton sequences in reverse order, which facilitates learning richer high-level semantics and better gait representations. Second, inspired by the fact that motion's continuity endows temporally adjacent skeletons with higher correlations (``\textit{locality}''), we propose a locality-aware attention mechanism that encourages learning larger attention weights for temporally adjacent skeletons when reconstructing current skeleton, so as to learn locality when encoding gait. Finally, we propose \textit{Attention-based Gait Encodings} (AGEs), which are built using context vectors learned by locality-aware attention, as final gait representations. AGEs are directly utilized to realize effective person Re-ID. Our approach typically improves existing skeleton-based methods by 10-20\% \textit{Rank-1} accuracy, and it achieves comparable or even superior performance to multi-modal methods with extra RGB or depth information\footnotemark[2].
\end{abstract}
\footnotetext[2]{Codes are available at \href{https://github.com/Kali-Hac/SGE-LA}{https://github.com/Kali-Hac/SGE-LA.}}
\section{Introduction}

\begin{figure}[t]
\begin{center}
\scalebox{0.38}{
  \includegraphics[]{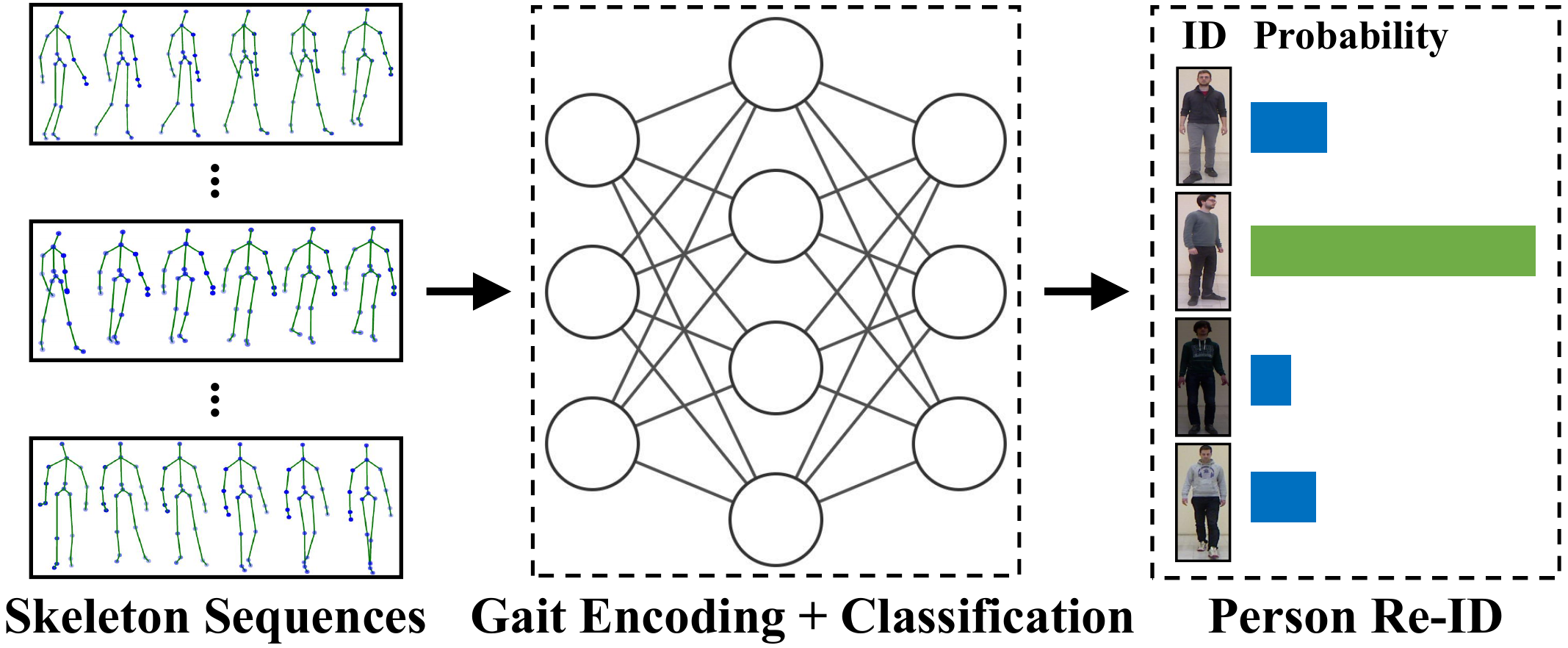}
  }
\end{center}
  \caption{Gait-based person Re-ID using 3D skeleton data.}
\label{figure1}
\end{figure}

Fast, effective and reliable person re-identification (Re-ID) is vital for various applications such as security authentication, human tracking, and role-based activity understanding \cite{vezzani2013people}. For person Re-ID, \textit{gait} is one of the most useful human body characteristics, and it has drawn increasing attention from researchers since it can be extracted by unobtrusive methods without co-operative subjects \cite{connor2018biometric}. Existing physiological and psychological studies \cite{murray1964walking,cutting1977recognizing} suggest that different human individuals have different gait characteristics, which contain numerous unique and relatively stable patterns ($e.g.$, stride length and angles of body joints). In the field of gait analysis, gait can be described by two types of methods: \textbf{(a)} \textit{Appearance-based methods} \cite{zhang2019comprehensive}, which exploit human silhouettes from aligned image sequences to depict gait. However, this type of methods are vulnerable to body shape changes and appearance changes. \textbf{(b)} \textit{Model-based methods} \cite{liao2020model}, which model gait by body structure and motion of human body joints. Unlike appearance-based methods, model-based methods are scale and view invariant \cite{nambiar2019gait}, thus leading to better robustness in practice. Among different models, 3D skeleton model, which describes humans by 3D coordinates of key body joints, is a highly efficient representation of human body structure and motion \cite{han2017space}. Compared with RGB or depth data, 3D skeleton enjoys many merits like better robustness and much smaller data size, and they can be easily collected by modern devices like Kinect. Therefore, exploiting 3D skeleton data to perform gait analysis for downstream tasks like person Re-ID (illustrated in Fig. \ref{figure1}) has gained surging popularity \cite{liao2020model}. Nevertheless, the way to extract or learn discriminative gait features from 3D skeleton sequence data remains open to be explored.

To this end, most existing works like \cite{barbosa2012re,andersson2015person} rely on hand-crafted skeleton descriptors. However, it is usually complicated and tedious to design such hand-crafted descriptors, $e.g.$, \cite{andersson2015person} defines 80 skeleton descriptors from the views of anthropometric and gait attributes for person Re-ID. Besides, they heavily rely on domain knowledge such as human anatomy \cite{yoo2002extracting}, thus lacking the ability to mine latent gait features beyond human cognition. To alleviate this problem, few recent works like \cite{haque2016recurrent} resort to deep neural networks to learn gait representations automatically. However, they all follow the classic \textit{supervised} learning paradigm and require labeled data, so they cannot perform gait representation learning with unlabeled skeleton data. In this paper, we for the first time propose a \textbf{\textit{self-supervised}} approach with \textbf{\textit{locality-aware}} attention mechanism, which only requires unlabeled 3D skeleton data for gait encoding. First, by introducing certain self-supervision as a high-level learning goal, we not only enable our model to learn gait representations from unlabeled skeleton data, but also encourage the model to capture richer high-level semantics ($e.g.$, sequence order, body part motion) and more discriminative gait features. Specifically, we propose a self-supervised learning objective that aims to reconstruct the input skeleton sequence in reverse order, which is implemented by an encoder-decoder architecture. Second, since the continuity of motion leads to very small pose/skeleton changes in a small time interval \cite{aggarwal1999human}, this endows skeleton sequences with the property of locality: For each skeleton frame in a skeleton sequence, its temporally adjacent skeleton frames within a local context enjoy higher correlations to itself. Thus, to enable better skeleton reconstruction and gait representation learning, we propose the locality-aware attention mechanism to incorporate such locality into the gait encoding process.
Last, we leverage the context vectors learned by the proposed locality-aware attention mechanism to construct Attention-based Gait Encodings (AGEs) as the final gait representations. We demonstrate that AGEs, which are learned without skeleton sequence labels, can be directly applied to person Re-ID and achieve highly competitive performance.

In summary, we make the following contributions:
\begin{itemize}
\item 
We propose a self-supervised approach based on reverse sequential skeleton reconstruction and an encoder-decoder model, which enable us to learn discriminative gait representations without skeleton sequence labels.

\item
We propose a locality-aware attention mechanism to exploit the locality nature in skeleton sequences for better skeleton reconstruction and gait encoding.

\item We propose AGEs as novel gait representations, which are shown to be highly effective for person Re-ID.

\end{itemize}

Experiments demonstrate the effectiveness of our approach in gait encoding/person Re-ID. It is also shown that our model is readily transferable among datasets, which validates its abilities to learn transferable high-level features of skeletons.

\section{Related Work} 
\label{related}
\textbf{Person Re-ID using Skeleton Data.} \
For person Re-ID, most existing works extract hand-crafted features to depict certain geometric, morphological or anthropometric attributes of 3D skeleton data.
\cite{barbosa2012re} computes 7 Euclidean distances between the floor plane and joint or joint pairs to construct a distance matrix, which is learned by a quasi-exhaustive strategy to perform person Re-ID. 
\cite{munaro2014one} further extends them to 13 skeleton descriptors ($D^{13}$) and leverages support vector machine (SVM) and k-nearest neighbor (KNN) for classification.
In \cite{pala2019enhanced}, 16 Euclidean distances between body joints ($D^{16}$) are fed to Adaboost to realize Re-ID.
Since skeleton based features alone are hard to achieve satisfactory Re-ID performance, features from other modalities ($e.g.$, 3D point clouds \cite{munaro20143d}, 3D face descriptor \cite{pala2019enhanced}) are also used to enhance the performance. Meanwhile, few recent works exploit supervised deep learning models to learn gait representations from skeleton data: 
\cite{haque2016recurrent} utilizes long short-term memory (LSTM) \cite{hochreiter1997long} to model temporal dynamics of body joints to perform person Re-ID; \cite{liao2020model} proposes PoseGait, which feeds 81 hand-crafted pose features of 3D skeleton data into convolutional neural networks (CNN) for human recognition.

\paragraph{Depth-based and Multi-modal Person Re-ID Methods.} Depth-based methods typically exploit human shapes or silhouettes from depth images to extract gait features for person Re-ID. \cite{sivapalan2011gait} extends the Gait Energy Image (GEI) \cite{chunli2010behavior} to 3D domain and proposes Gait Energy Volume (GEV) algorithm based on depth images to perform gait-based human recognition. 3D point clouds based on depth data are also widely used to estimate body shape and motion trajectories. \cite{munaro2014one} proposes point cloud matching (PCM) to compute the distances of multi-view point cloud sets, so as to discriminate different persons. \cite{haque2016recurrent} adopts 3D LSTM to model motion dynamics of 3D point clouds for person Re-ID. As to multi-modal methods, they usually combine skeleton-based features with extra RGB or depth information ($e.g.,$ depth shape features based on point clouds \cite{munaro20143d,hasan2016long,wu2017robust}) to boost Re-ID performance. In \cite{karianakis2018reinforced}, CNN-LSTM with reinforced temporal attention (RTA) is proposed for person Re-ID based on a split-rate RGB-depth transfer approach. 

\section{The Proposed Approach}
\label{proposed}

\begin{figure*}[t]
    \centering
    \scalebox{0.65}{
    \includegraphics{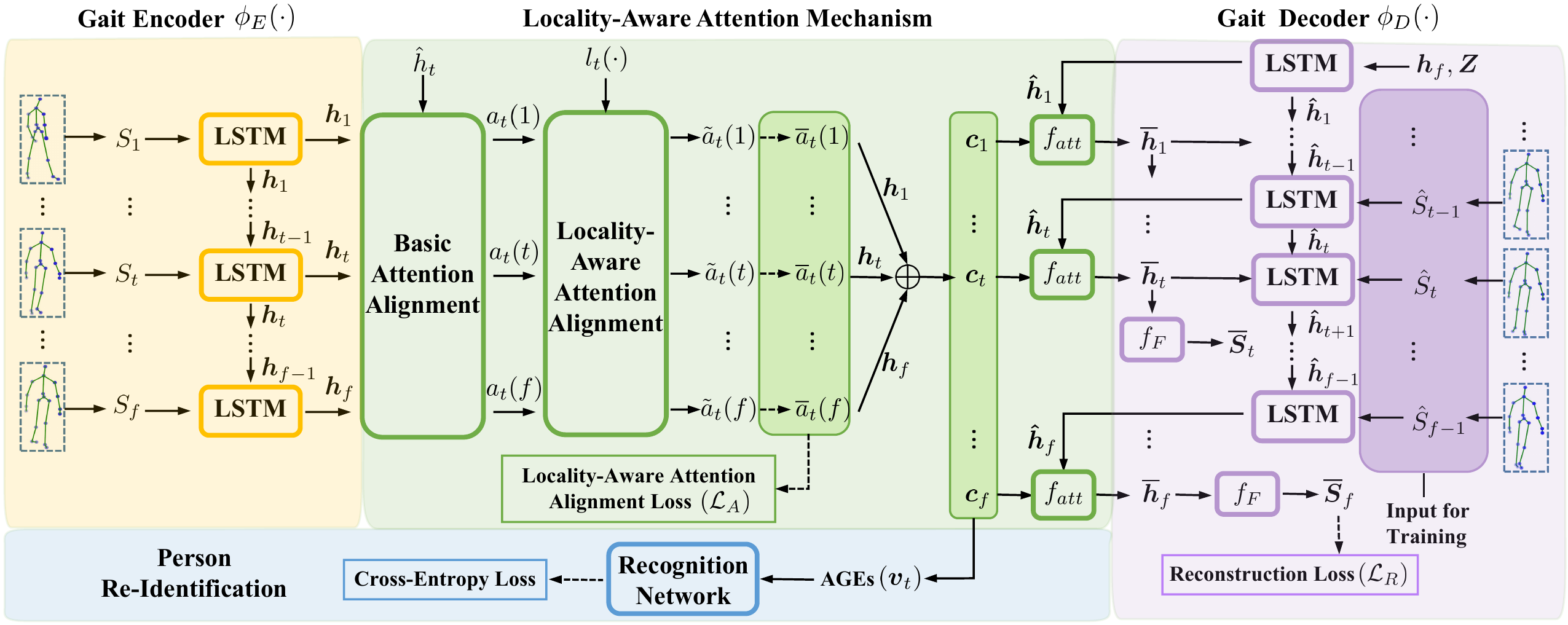}
    }
    \caption{Flow diagram of our model: (1) Gait Encoder (yellow) encodes each skeleton frame $S_t$ into an encoded gait state $\boldsymbol{h}_t$. (2) Locality-aware attention mechanism (green) first computes the basic attention alignment score ${a}_{t}(\cdot)$, so as to measure the content-based correlation between each encoded gait state and the decoded gait state $\boldsymbol{{\hat h}}_{t}$ from Gait Decoder (purple). Then, the locality mask $l_{t}(\cdot)$ provides an objective $\tilde{a}_{t}(\cdot)\!=\!{a}_{t}(\cdot)\ l_{t}(\cdot)$, which guides our model to learn locality-aware alignment scores $\overline{a}_{t}(\cdot)$ by the locality-aware attention alignment loss ${\mathcal L}_{A}$. 
    Next, $\boldsymbol{h}_1\cdots \boldsymbol{h}_f$ are weighted by $\overline{a}_{t}(\cdot)$ to compute the context vector $\boldsymbol{c}_{t}$. $\boldsymbol{c}_{t}$ and $\boldsymbol{{\hat h}}_{t}$ are fed into the concatenation layer $f_{att}(\cdot)$ to produce an attentional state vector $\boldsymbol{\overline{h}}_{t}$.
    Finally, $\boldsymbol{\overline{h}}_{t}$ is fed into the full connected layer $f_{F}(\cdot)$ to predict $t^{th}$ skeleton $\boldsymbol{\overline{S}}_{t}$ and Gait Decoder for later decoding. (3) $\boldsymbol{c}_{t}$ is used to build Attention-based Gait Encodings (AGEs) $\boldsymbol{v}_{t}$, which are fed into a recognition network for person Re-ID (blue).}
\label{model}
\end{figure*}
Suppose that an input skeleton sequence $\boldsymbol{S}=(S_{1},\cdots,S_{f})$ contains $f$ consecutive skeleton frames, where  $S_{i} \in \mathbb{R}^{J \times 3}$ contains 3D coordinates of $J$ body joints. The training set $\Phi=\{\boldsymbol{S}^{(i)}\}_{i=1}^{N}$ contains $N$ skeleton sequences collected from different persons. Each skeleton sequence $\boldsymbol{S}^{(i)}$ corresponds to a label $y_i$, where $ \ y_i \!\in\! \left \{ 1,\cdots, C \right \}$ and $C$ is the number of persons. Our goal is to learn discriminative gait features $\boldsymbol{v}$ from $\boldsymbol{S}$ without using any label. Then, the effectiveness of learned features $\boldsymbol{v}$ is validated by using them to perform person Re-ID: Learned features and labels are used to train a simple recognition network (note that the learned features $\boldsymbol{v}$ are fixed and NOT tuned by training at this stage). The overview of the proposed approach is given in Fig. \ref{model}, and we present the details of each technical component below. 

\subsection{Self-Supervised Skeleton Reconstruction}
\label{reconstruction_mechanism}

To learn gait representations without labels, we propose to introduce self-supervision by learning to reconstruct input skeleton sequences in \textit{reverse} order, $i.e.$, by taking $\boldsymbol{{ S}}$ as inputs, we expect our model to output the sequence $\boldsymbol{{\hat S}}=({\hat S}_{1},\cdots,{\hat S}_{f})=(S_{f},\cdots,S_{1})$, which gives ${\hat S}_{t}=S_{f-t+1}$. Compared with na\"ive reconstruction that learns to reconstruct exact inputs ($\boldsymbol{S}\!\longmapsto\! \boldsymbol{S}$), the proposed learning objective ($\boldsymbol{S}\!\longmapsto\! \boldsymbol{{\hat S}}$) is endowed with high-level information (skeleton order in a sequence) that is meaningful to human perception, which requires the model to capture richer high-level semantics (\textit{e.g.}, body parts, motion patterns) to achieve this learning objective. In this way, our model is expected to learn better gait representations than frequently-used plain reconstruction. Formally, given an input skeleton sequence, we use the encoder to encode each skeleton frame $S_{t}$ ($t\in\{1,\cdots f\}$) and the previous step's latent state $\boldsymbol{h}_{t-1}$ (if existed), which provides context information, into current latent state $\boldsymbol{h}_{t}$:
\begin{align*}
\boldsymbol{h}_{t}=\left\{
\begin{aligned}
\ & \phi_{E}\left(S_{1} \right) \quad & \text{if} \quad t = 1 \\
\ & \phi_{E}\left(\boldsymbol{h}_{t-1}, S_{t} \right) \quad & \text{if} \quad t > 1
\end{aligned}
\right. \tag{$1$}\label{encode}
\end{align*}
where $\boldsymbol{h}_{t} \in \mathbb{R}^{k}$, $\phi_{E}(\cdot)$ denotes our Gait Encoder (GE). GE is built with an LSTM, which aims to capture long-term temporal dynamics of skeleton sequences. $\boldsymbol{h}_{1},\cdots,\boldsymbol{h}_{f}$ are \textit{encoded gait states} that contain preliminary gait encoding information. In the \textbf{training} phase, encoded gait states are decoded by a Gait Decoder (GD) to reconstruct the target sequence $\boldsymbol{{\hat S}}$, and the decoding process is performed below (see  Fig. \ref{model}):
\begin{align*}
(\boldsymbol{{\hat h}}_{t}, \overline{S}_{t})=\left\{
\begin{aligned}
\ & \phi_{D}\left(\boldsymbol{h}_{f}, \boldsymbol{Z}\right) \quad & \text{if} \quad t = 1 \\
\ & \phi_{D}\left(\boldsymbol{{\hat h}}_{t-1}, {\hat S}_{t-1},\boldsymbol{\overline{h}}_{t-1}\right) \quad & \text{if} \quad t > 1
\end{aligned}
\right.\tag{$2$}\label{decode_train}
\end{align*}
where $\phi_{D}(\cdot)$ denotes the GD. GD consists of an LSTM and a fully connected (FC) layer that outputs predicted skeletons. $ \boldsymbol{{\hat h}}_{t}\!\in \mathbb{R}^{k}$ refers to the $t^{th}$ \textit{decoded gait state}, $i.e.$, the latent state output by GD's LSTM to predict $t^{th}$ skeleton $\overline{S}_t$. When the decoding is initialized ($t=1$), we feed an all-0 skeleton placeholder $\boldsymbol{Z}\in \mathbb{R}^{J}$ and the final encoded gait state $\boldsymbol{h}_f$ into GD to decode the first skeleton. Afterwards, to predict $t^{th}$ skeleton $\overline{S}_t$, $\phi_{D}(\cdot)$ takes three inputs from the ${(t-1)}^{th}$ decoding step: decoded gait state $\boldsymbol{{\hat h}}_{t-1}$, the ground truth skeleton $\hat{S}_{t-1}$ ($S_{f-t+2}$) that enables better convergence, and the \textit{attentional state vector} $\boldsymbol{\overline{h}}_{t-1}$ that fuses encoding and decoding information based on the proposed attention mechanism, which will be elaborated in Sec. \ref{sec:attention}. In this way, we define the objective function $\mathcal{L}_{R}$ for skeleton reconstruction, which minimizes the mean square error (MSE) between a target skeleton sequence $\boldsymbol{{\hat S}}$ and a predicted skeleton sequence $\boldsymbol{\overline{S}}$: 
\begin{align*}
\mathcal{L}_{R}=\sum_{i=1}^{f}\sum_{j=1}^{J}(\boldsymbol{\overline{S}}_{ij}-\boldsymbol{{\hat S}}_{ij})^{2} \tag{$3$}\label{reconstruct}
\end{align*}
where $\boldsymbol{\overline{S}}_{ij}$, $\boldsymbol{{\hat S}}_{ij}$ represent the $j^{th}$ joint position of the $i^{th}$ predicted or target skeleton. In the \textbf{testing} phase, to test the reconstruction ability of our model, we use the predicted skeleton $\overline{S}_{t-1}$ rather than target skeleton $\hat S_{t-1}$ as the input to $\phi_{D}$ in the $t>1$ case, $i.e.$,  $(\boldsymbol{{\hat h}}_{t},\overline{S}_{t}) \!=\!\phi_{D}\left( \boldsymbol{{\hat h}}_{t-1}, \overline{S}_{t-1},\boldsymbol{\overline{h}}_{t-1} \right)$. To facilitate training, our implementation actually optimizes Eq. \ref{reconstruct} on each individual dimension of the skeleton's 3D coordinates: $\boldsymbol{S}^{[d]}{\longmapsto}\boldsymbol{{\hat S}}^{[d]}$, where $d\in\{X, Y, Z\}$ corresponds to a certain dimension of 3D space, and $\boldsymbol{S}^{[d]},\boldsymbol{{\hat S}}^{[d]}\in \mathbb{R}^{f \times J}$.

\subsection{Locality-Aware Attention Mechanism}
\label{sec:attention}

As learning gait features requires capturing motion patterns from skeleton sequences, it is natural to consider a straightforward property of motion---Continuity. Motion's continuity ensures that those skeletons in a small temporal interval will NOT undergo drastic changes, thus resulting in higher inter-skeleton correlations in this local temporal interval, which is referred as ``\textit{\textbf{locality}}''. Due to such locality, when reconstructing one skeleton in a sequence, we expect our model to pay more attention to its neighboring skeletons in the local context. 
To this end, we propose a novel locality-aware attention mechanism, with its modules presented below:

 \begin{figure}[t]
    \centering
     \subfigure[Attention Matrix]{\scalebox{0.34}{\label{fig:attention_matrix}\includegraphics[]{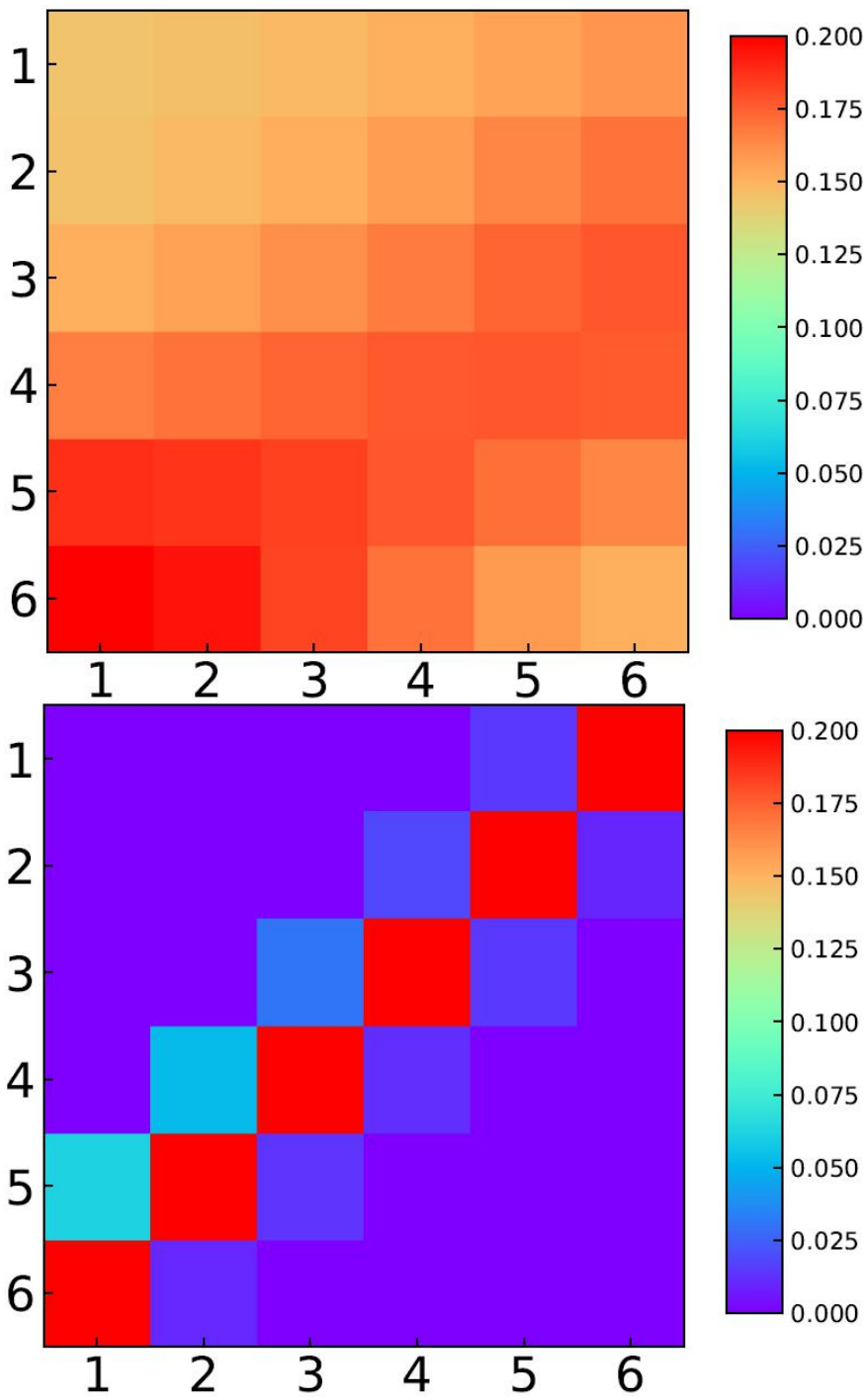}}
     }
    \subfigure[Reconstruction Loss Curves]{\scalebox{0.33}{\label{fig:reconstruction_loss}\includegraphics[]{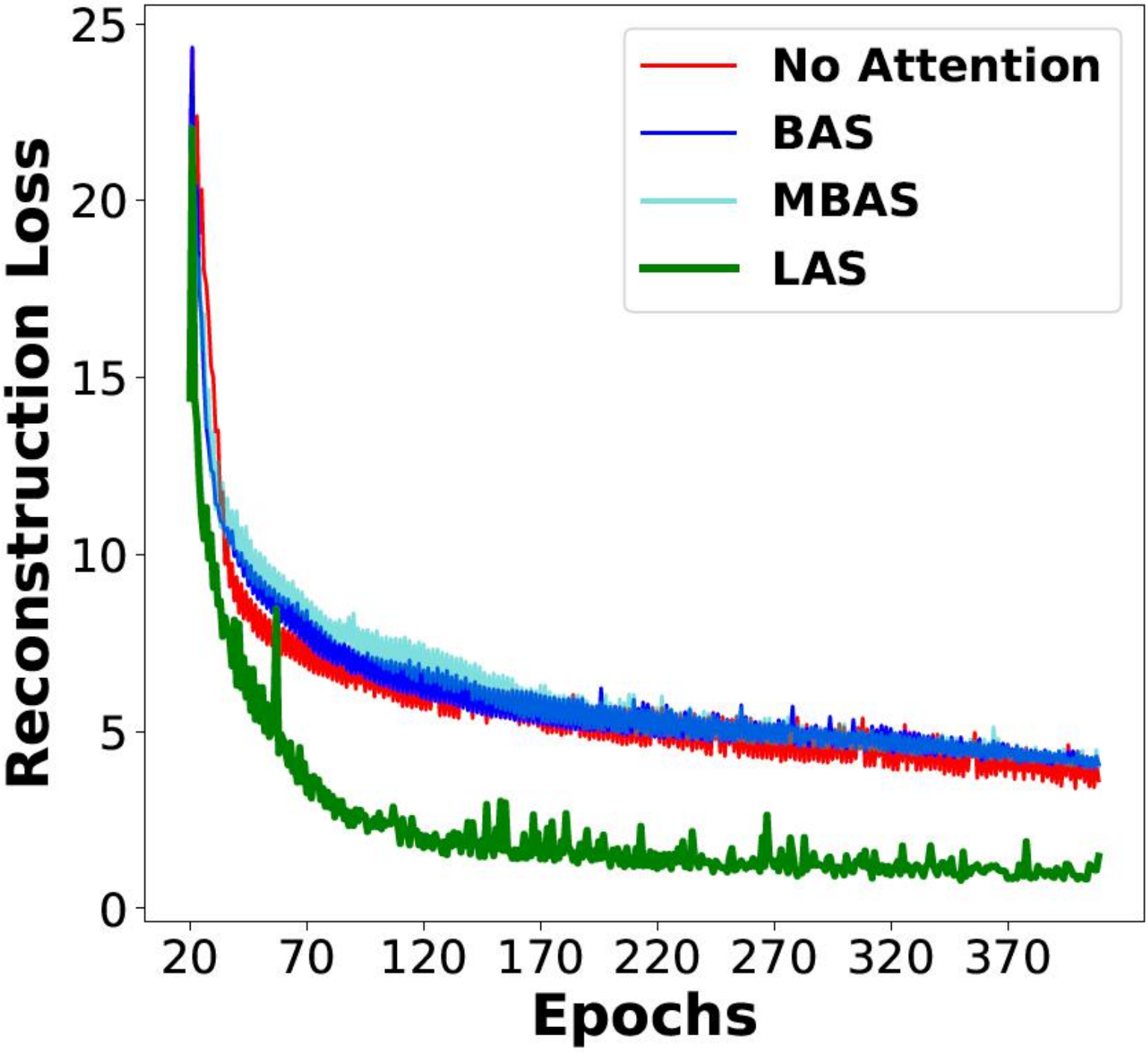}}
    }
    \caption{(a) Visualization of the BAS (top) and LAS (bottom) attention matrices that represent average attention alignment scores. Note that the abscissa and ordinate denote indices of input skeletons and predicted skeletons respectively. (b) Reconstruction loss curves when using no attention, BAS, MBAS or LAS for reconstruction. }
    \label{attention_comparison}
\end{figure}

\textbf{Basic Attention Alignment.} We first introduce Basic Attention (BA) alignment \cite{luong-etal-2015-effective} to measure the content ($i.e.$, latent state) based correlations between the input sequence and the predicted sequence. As shown in Fig. \ref{model}, at $t^{th}$ decoding step, we compute the \textit{BA Alignment Scores (BAS)} between $\boldsymbol{{\hat h}}_{t}$ and encoded gait state $\boldsymbol{h}_{j}$ ($j\in\{1,\dots,f\}$):
\begin{align*}
     {a}_{t}(j)&=\operatorname{align}\left(\boldsymbol{{\hat h}}_{t},\boldsymbol{h}_{j}\right)=\frac{\exp \left({\boldsymbol{{\hat h}}_{t}}^{\top}\boldsymbol{h}_{j}\right)}{\sum^{f}_{i=1} \exp \left({\boldsymbol{{\hat h}}_{t}}^{\top}\boldsymbol{h}_{i}\right)}
    \tag{$4$}\label{align}
 \end{align*}
 BAS aims to focus on those more correlative skeletons in the encoding stage, and provide preliminary attention weights for skeleton decoding. However, BA alignment only considers the content based correlations and does not explicitly take locality into consideration, which motivates us to propose locality mask and locality-aware attention alignment below.

\textbf{Locality Mask.} Our motivation is to incorporate locality into the gait encoding process for better skeleton reconstruction. As the goal is to decode the $t^{th}$ skeleton $\overline{S}_t$ as $\hat S_t$, we assume those skeletons in the local temporal context of $S_{p_{t}}$ to be highly correlated to $\overline{S}_t$, where $p_{t}=f-t+1$ (note that we use reverse reconstruction). To describe the local context centered at $S_{p_{t}}$, we define an attentional window $[p_{t}-D, p_{t}+D]$, where $D$ is a selected integer to control the attentional range. Since the locality will favor temporal positions near $p_{t}$ (closer positions are more correlative), a direct solution is to place a \textit{Gaussian} distribution centered around $p_{t}$ as a locality mask:
\begin{align*}
{l}_{t}(j)&=\mathrm{exp}\left ( -\frac{(j-p_{t})^2}{2\sigma^{2}} \right )
\tag{$5$}\label{local_align}
\end{align*}
where we empirically set $\sigma \!= \!\frac{D} {2}$, $j$ is a position within the window centered at $p_{t}$. We can weight BAS by this locality mask to compute \textit{Masked BA Alignment Scores (MBAS)} below, which directly forces alignment scores to obtain locality:
\begin{align*}
\tilde{a}_{t}(j)={l}_{t}(j)\cdot{a}_{t}(j)
\tag{$6$}\label{local_align_scores}
\end{align*}

\textbf{Locality-Aware Attention Alignment.} Despite that the locality mask is straightforward to yield locality, it is a very coarse solution that brutally changes the alignment scores. Therefore, instead of using MBAS (${\tilde{a}}_{t}(j)$) directly, we propose the \textit{Locality-aware Attention (LA) alignment}. Specifically, an LA alignment loss term $\mathcal{L}_A$ is used to encourage LA alignment to learn similar locality of the ${\tilde{a}}_{t}(j)$:
\begin{align*}
\mathcal{L}_{A}=\sum_{t=1}^{f}\sum_{j=1}^{f}({a}_{t}(j)-\tilde{a}_{t}(j))^{2} \tag{$7$}\label{LA_loss}
\end{align*}
 By adding the loss term $\mathcal{L}_{A}$, we can obtain \textit{LA Alignment Scores (LAS)}. Note that in Eq. \ref{align}, the final learned ${a}_{t}(j)$ is BAS. For clarity, we use $\overline{a}_{t}(j)$ to represent learned LAS here.
 
With the guidance of $\mathcal{L}_{A}$, our model learns to allocate more attention to the local context by itself, rather than using a locality mask. To utilize alignment scores to yield an attention-weighted encoded gait state at the $t^{th}$ step, we calculate the \textit{context vector} $\boldsymbol{c}_t$ by a sum of weighted encoded gait states:
\begin{align*}
\boldsymbol{c}_{t}=\sum^{f}_{j=1} \overline{a}_{t}(j)\boldsymbol{h}_{j} \tag{$8$}\label{context}
\end{align*}
Note that context vector $\boldsymbol{c}_{t}$ can also be computed by BAS or MBAS. $\boldsymbol{c}_{t}$ provides a synthesized gait encoding that is more relevant to $\boldsymbol{\hat{h}}_{t}$, which facilitates the reconstruction of $t^{th}$ skeleton.  To combine both encoding and decoding information for reconstruction, we use a concatenation layer $f_{att}(\cdot)$ that combines $\boldsymbol{c}_{t}$ and $\boldsymbol{{\hat h}}_{t}$ into an attentional state vector $\boldsymbol{\overline{h}}_{t}$: 
\begin{align*}
     \boldsymbol{\overline{h}}_{t}&=f_{att}(\boldsymbol{c}_{t}, \boldsymbol{{\hat h}}_{t})=\tanh (\boldsymbol{W}_{att}[\boldsymbol{c}_{t};\boldsymbol{{\hat h}}_{t}]) \tag{$9$}\label{attention_state}
\end{align*}
where $\boldsymbol{W}_{att}$ represents the learnable weight matrix in the layer. 
Finally, we generate (reconstruct) the $t^{th}$ skeleton by the FC layer $f_F(\cdot)$ of the GD:
\begin{align*}
         \overline{S}_{t}=f_{F}\left(\boldsymbol{\overline{h}}_{t}\right)=\boldsymbol{W}_{F}\boldsymbol{\overline{h}}_{t} \tag{$10$}\label{prediction}
\end{align*}
where $\boldsymbol{W}_{F}$ is the weights to be learned in this FC layer.

\textbf{Remarks:}   
\textbf{(a)} To provide an intuitive illustration of the proposed locality-aware attention mechanism, we visualize the BAS and LAS attention matrices, which are formed by average alignment scores computed on dimension $X$ as an example. As shown by Fig. \ref{fig:attention_matrix}, LA alignment significantly improves the locality of learned alignment scores: Relatively large alignment scores are densely distributed near the clinodiagonal line of the alignment score matrix (clinodiagonal line reflects skeletons' correlations to themselves when performing reverse reconstruction), which means adjacent skeletons are assigned with larger attention than remote skeletons. By contrast, despite that BA alignment also learns locality to some extent, BA's alignment weights show a much more even distribution, and many non-adjacent skeletons are also given large alignment scores. Similar trends are also observed for dimension $Y$ and $Z$. \textbf{(b)} The proposed LA alignment enables us to learn better reconstruction of skeleton sequences. We visualize the reconstruction loss during training in four cases: using no attention mechanism, BAS, MBAS and LAS. As shown by Fig. \ref{fig:reconstruction_loss}, it can be observed that training with LAS converges at a faster speed with an evidently smaller reconstruction loss, which justifies our intuition that locality will facilitate training. Interestingly, we observe that using the locality mask directly in fact does not benefit training, and it verifies that learning is a better way to obtain locality.

\subsection{Attention-based Gait Encodings (AGEs)}
\label{GAEV_construction}

Since our ultimate goal is to learn good gait features from skeleton data to perform person Re-ID, we need to extract certain internal skeleton embedding from the proposed model as gait representations. Unlike traditional LSTM based methods that basically rely on the last hidden state to compress the temporal dynamics of a sequence \cite{weston2014memory}, we recall that the dynamic context vector $\mathbf{c}_t$ from the attention mechanism integrates the key encoded gait states of input skeletons and retains crucial spatio-temporal information to reconstruct target skeletons. Hence, we utilize them instead of the last hidden state to build our final gait representations---AGEs. A skeleton-level AGE ($\boldsymbol{v}_t$) is defined as follows:
\begin{align*}
\boldsymbol{v}_{t}=[\boldsymbol{c}^{X}_{t};\boldsymbol{c}^{Y}_{t};\boldsymbol{c}^{Z}_{t}]
\tag{$11$}\label{eq9}
\end{align*}
where $\boldsymbol{c}^{d}_{t}$ denotes the context vector computed on dimension $d\in\{X, Y, Z\}$ of the $t^{th}$ step in decoding. To perform person Re-ID, we use $\boldsymbol{v}_{t}$ to train a simple recognition network $f_{RN}(\cdot)$ that consists of a hidden layer and a softmax layer. We average the prediction of each skeleton-level AGE $f_{RN}(\boldsymbol{v}_{t};\theta_{\tau})$ ($t\in\{1,\cdots,f\}$) in a skeleton sequence to be the final sequence-level prediction for person Re-ID, where $\theta_{\tau}$ refers to parameters of $f_{RN}(\cdot)$. Note that during training, each skeleton in one sequence shares the same skeleton sequence label $y_i$. Besides, skeleton labels are only used to train the recognition network, $i.e.$, AGEs are fixed during training to demonstrate the effectiveness of learned gait features.  

\subsection{The Entire Approach}
\label{optimization}
As a summary, the computation flow of the entire approach during skeleton reconstruction is $\boldsymbol{h}\rightarrow\boldsymbol{\hat {h}}\rightarrow \overline{a}\rightarrow \boldsymbol{c}\rightarrow \boldsymbol{\overline{h}}\rightarrow \boldsymbol{\overline{S}}$. To guide model training in the reconstruction process, we combine the skeleton reconstruction loss $\mathcal{L}_{R}$ in Eq. \ref{reconstruct} and the LA alignment loss $\mathcal{L}_{A}$ in Eq. \ref{LA_loss} as follows:
\begin{align*}
&\mathcal{L}=\lambda_{R} \mathcal{L}_{R}+\lambda_{A}\mathcal{L}_{A}+\beta \left\|\Theta\right\|^{2}_{2} \tag{$12$}\label{eq11}
\end{align*}
where $\Theta$ denotes the parameters of the model,  $\lambda_{R}$, $\lambda_{A}$ are the weight coefficients to trade off the importance of the reconstruction loss and LA alignment loss,  $\left\|\Theta\right\|^{2}_{2}$ is $L_{2}$ regularization.
For the person Re-ID task, we employ the cross-entropy loss to train the recognition network with AGEs ($\boldsymbol{v}_t$).

\section{Experiments}
\label{experiments}
\subsection{Experimental Settings}
\quad\textbf{Datasets:} We evaluate our method on three public Re-ID datasets that provide 3D skeleton data: BIWI \cite{munaro2014one}, IAS-Lab \cite{munaro2014feature} and Kinect Gait Biometry Dataset (KGBD) \cite{andersson2015person}. They collect skeleton data from 50, 11 and 164 different individuals respectively. We follow the evaluation setup in \cite{haque2016recurrent}, which is frequently used in the literature: For BIWI, we use the full training set and the Walking testing set that contains dynamic skeleton data; For IAS-Lab, we use the full training set and two test splits, IAS-A and IAS-B; For KGBD, since no training and testing splits are given, we randomly leave one skeleton video of each person for testing and use the rest of videos for training. The experiments are repeated for multiple times and the average performance is reported on KGBD. We discard the first and last 10 frames of each original skeleton sequence to avoid ineffective skeleton recording. Then, we spilt the training dataset into multiple skeleton sequences with the length $f$, and two consecutive sequences share $\frac{f}{2}$ overlapping skeletons, which aims to obtain as many skeleton sequences as possible to train our model.

\textbf{Implementation Details:} The number of body joints is set to the maximum number in all datasets, namely $J=20$. The sequence length $f$ is empirically set to 6 as it achieves the best performance in average among different sequence length settings. To learn the locality-aware attention for the whole sequence, the attentional range $D$ of LA is set to 6. We use a 2-layer LSTM with $k = 256$ hidden units per layer for both GE and GD. We empirically set both  $\lambda_{R}$ and $\lambda_{A}$ to 1, while a momentum 0.9 is utilized for optimization. We use a learning rate $lr=0.0005$, and we set the weight of $L_{2}$ regularization $\beta$ to 0.02. The batch size is set to 128 in all experiments. 

\textbf{Evaluation Metrics:} 
Person Re-ID typically follows a ``multi-shot'' manner that leverages predictions of multiple frames or a sequence-level representation to produce a sequence label. In this paper, we compute both \textit{Rank-1} accuracy and \textit{nAUC} (area under the cumulative matching curve (CMC) normalized by the number of ranks \cite{gray2008viewpoint}) to evaluate multi-shot person Re-ID performance.

\begin{table*}[ht]
\centering
\scalebox{0.9}{
\begin{tabular}{ll|rrrr|rrrr}
\specialrule{0.1em}{0.45pt}{0.45pt}
 &  & \multicolumn{4}{c|}{\textbf{\textit{Rank-1 (\%)}}} & \multicolumn{4}{c}{\textbf{\textit{nAUC}}} \\ \specialrule{0.1em}{0.2pt}{0.2pt}
\textbf{Id} & \textbf{Methods} & \textbf{BIWI} & \textbf{IAS-A} & \multicolumn{1}{l}{\textbf{IAS-B}} & \textbf{KGBD} & \textbf{BIWI} & \textbf{IAS-A} & \multicolumn{1}{l}{\textbf{IAS-B}} & \textbf{KGBD} \\ \specialrule{0.1em}{0.45pt}{0.45pt}
1 & Gait Energy Image \shortcite{chunli2010behavior} & 21.4 & 25.6 & 15.9 & —  & 73.2 & 72.1 & 66.0 & — \\
2 & Gait Energy Volume \shortcite{sivapalan2011gait} & 25.7 & 20.4 & 13.7 & — & 83.2 & 66.2 & 64.8 & — \\
3 & 3D LSTM \shortcite{haque2016recurrent} & 27.0 & 31.0 & 33.8 & — & 83.3 & 77.6 & 78.0 & — \\
\specialrule{0.1em}{0.45pt}{0.45pt}
4 & PCM + Skeleton \shortcite{munaro20143d}  & 42.9 & 27.3 & 81.8 & — & — & — & — & —\\
5 & Size-Shape decriptors + SVM \shortcite{hasan2016long}  & 20.5 & — & — & — & — & — & — & —\\
6 & Size-Shape decriptors + LDA \shortcite{hasan2016long}  & 22.1 & — & — & — & — & — & — & — \\
7 & DVCov + SKL \shortcite{wu2017robust} & 21.4 & 46.6 & 45.9 & — & — & — & — & — \\
8 & ED + SKL \shortcite{wu2017robust} & 30.0 & 52.3 & 63.3 & — & — & — & — & — \\
9 & CNN-LSTM with RTA \shortcite{karianakis2018reinforced}  & 50.0 & — & — & — & — & — & — & —\\
\specialrule{0.1em}{0.45pt}{0.45pt}
10 & $D^{13}$ descriptors + SVM \shortcite{munaro2014one}  & 17.9 & — & — & — & — & — & —  & —\\
11 & $D^{13}$ descriptors + KNN \shortcite{munaro2014one}   & 39.3 & 33.8 & 40.5 & 46.9  & 64.3 & 63.6 & 71.1  & 90.0\\
12 & $D^{16}$ descriptors + Adaboost \shortcite{pala2019enhanced}  & 41.8 & 27.4 & 39.2 & 69.9  & 74.1 & 65.5 & 78.2 & 90.6 \\
13 & Single-layer LSTM \shortcite{haque2016recurrent}  & 15.8 & 20.0 & 19.1 & 39.8 & 65.8 & 65.9 & 68.4  & 87.2 \\
14 & Multi-layer LSTM \shortcite{zheng2019relational}  & 36.1 & 34.4 & 30.9 & 46.2  & 75.6 & 72.1 & 71.9 & 89.8 \\
15 & PoseGait \shortcite{liao2020model} & 33.3 & 41.4 & 37.1 & \textbf{90.6} & 81.8 & 79.9 & 74.8 & \textbf{97.8}
\\
\specialrule{0.1em}{0.45pt}{0.45pt}
16 & \textbf{Ours}  & \textbf{59.1} & \textbf{56.1} & \textbf{58.2} & 87.7 & \textbf{86.5} & \textbf{81.7} & \textbf{85.3} & 96.3  \\ \specialrule{0.1em}{0.2pt}{0.2pt}
\end{tabular}
}
\caption{Comparison with existing skeleton-based methods (10-15). Depth-based methods (1-3) and multi-modal methods (4-9) are also included as a reference. Bold numbers refer to the best performers among skeleton-based methods. ``—'' indicates no published result.}
\label{skeleton_results}
\end{table*}

\subsection{Performance Comparison}
\label{64}
In Table \ref{skeleton_results}, we conduct an extensive comparison with existing skeleton based person Re-ID methods (Id = 10-15) in the literature. In the meantime, we also include classic depth-based methods (Id = 1-3) and representative multi-modal methods (Id = 4-9) as a reference. We obtain observations below:

\textbf{Comparison with Skeleton-based Methods: } As shown by Table \ref{skeleton_results}, our approach enjoys obvious advantages over existing skeleton-based methods in terms of person Re-ID performance: First, our approach evidently outperforms those methods that rely on manually-designed geometric or anthropometric skeleton descriptors (Id = 10-12). For example, $D^{13}$ (Id = 11) and recent $D^{16}$ (Id = 12) are two most representative hand-crafted feature based methods, and our model outperforms both of them by a large margin ($17.3\%$-$40.8\%$ \textit{Rank-1} accuracy and $5.7\%$-$22.2\%$ \textit{nAUC} on different datasets). Second, our approach is also superior to recent skeleton based methods that utilize deep neural networks (Id = 13-15). Our approach is the best performer on three out of four datasets (BIWI, IAS-A, IAS-B) with a $14.7\%$-$43.3\%$ \textit{Rank-1} accuracy and $1.8\%$-$20.7\%$ \textit{nAUC} gain. On KGBD dataset, our approach ranks $2^{nd}$ and performs slightly inferior to the latest PoseGait (Id = 15). However, despite that CNN is used, PoseGait still requires extracting 81 hand-crafted features. Besides, labeled skeleton data are indispensable for existing deep learning based methods, while our approach can learn better gait representations by unlabeled skeletons only.

\textbf{Comparison with Depth-based Methods and Multi-modal Methods:} 
\label{multi_approaches} Despite that our approach only takes skeleton data as inputs, our approach consistently outperforms baselines of classic depth-based methods (Id = 1-3) by at least $24\%$ \textit{Rank-1} and $3.2\%$ \textit{nAUC} gain. Considering the fact that skeleton data are of much smaller size than depth image data, our approach is both effective and efficient. As to the comparison with recent methods that exploit multi-modal inputs (Id = 4-9), the performance of our approach is still highly competitive: Although few multi-modal methods perform better on IAS-B, our skeleton based method achieves the best \textit{Rank-1} accuracy on BIWI and IAS-A. Interestingly, we note that the multi-modal approach that uses both point cloud matching (PCM) and skeletons yields the best accuray on IAS-B, but it performs markedly worse on datasets that undergo more frequent shape and appearance changes (IAS-A and BIWI). By contrast, our approach consistently achieves stable and satisfactory performance on each dataset. Thus, with 3D skeletons as the sole input, our approach can be a promising solution to person Re-ID and other potential skeleton-related tasks.

\begin{table}[t]
    \centering
    \scalebox{0.79}{
\begin{tabular}{ccccccc|c|c}
    \specialrule{0.1em}{0.45pt}{0.45pt}
\multicolumn{1}{l}{\textbf{GE}} & \multicolumn{1}{l}{\textbf{GD}} & \multicolumn{1}{l}{\textbf{Rev.}} & \multicolumn{1}{l}{\textbf{BAS}} & \multicolumn{1}{l}{\textbf{MBAS}} & \multicolumn{1}{l}{\textbf{LAS}} & \multicolumn{1}{l|}{\textbf{AGEs}} & \multicolumn{1}{l|}{\textbf{\textit{Rank-1}}} & \multicolumn{1}{l}{\textbf{\textit{nAUC}}} \\ \specialrule{0.1em}{0.2pt}{0.2pt}
\checkmark                      &                      &                        &                       &                        &                       &                         & 36.1                                 & 75.6                              \\
\checkmark                      & \checkmark                      &                        &                       &                        &                       &                         & 41.5                                 & 80.1                              \\
\checkmark                      & \checkmark                      & \checkmark                        &                       &                        &                       &                         & 46.7                                 & 81.5                              \\
\checkmark                      & \checkmark                      &                        & \checkmark                       &                        &                       & \checkmark                         & 45.7                                 & 84.1                             \\
\checkmark                      & \checkmark                      & \checkmark                        & \checkmark                       &                        &                       &                         &   53.3                               & 84.6                                  \\
\checkmark                      & \checkmark                      & \checkmark                        & \checkmark                       &                        &                       & \checkmark                         & 55.1                                 & 85.2                              \\
\checkmark                      & \checkmark                      & \checkmark                        &                       & \checkmark                        &                       &                         &  52.9                                    & 85.0                                  \\
\checkmark                               & \checkmark                               &                                   &                                  & \checkmark                                 &                                  & \checkmark                                  & 53.1                                & 83.6    \\
\checkmark                      & \checkmark                      & \checkmark                        &                       & \checkmark                        &                       & \checkmark                         & 54.5                                 & 85.6                              \\ 
\checkmark                      & \checkmark                      & \checkmark                        &                       &                        & \checkmark                       &                         & 57.7                                     &  85.8                                 \\ 
\checkmark                               & \checkmark                               &                                   &                                  &                                   & \checkmark                                & \checkmark                                  & 57.2                                & 85.7        \\
\checkmark                      & \checkmark                      & \checkmark                        &                       &                        & \checkmark                       & \checkmark                         & \textbf{59.1}                        & \textbf{86.5}                     \\ \specialrule{0.1em}{0.2pt}{0.2pt}
\end{tabular}
}
 \caption{Ablation study of our model. ``\checkmark'' \ indicates that the corresponding model component is used: GE, GD, reverse skeleton reconstruction (Rev.), different types of attention alignment scores (BAS, MBAS, LAS). ``AGEs'' indicates exploiting AGEs ($\boldsymbol{v}_{t}$) rather than encoded gait states of GE's LSTM $\boldsymbol{h}_{t}$ for person Re-ID.}
 \label{ablation_t}
\end{table}

\section{Discussion}
\label{discuss}
\quad\textbf{Ablation Study.}
We perform ablation study to verify the effectiveness of each model component. As shown in Table \ref{ablation_t}, we draw the following conclusions: \textbf{(a)} The proposed encoder-decoder architecture (GE-GD) performs remarkably better ($5.4\%$ \textit{Rank-1} accuracy and $4.5\%$ \textit{nAUC} gain) than supervised learning paradigm that uses GE only, which verifies the necessity of encoder-decoder architecture and skeleton reconstruction mechanism. \textbf{(b)} Using reverse reconstruction (Rev. in Table \ref{ablation_t}) produces evident performance gain ($1.4\%$-$9.4\%$ \textit{Rank-1} accuracy and $0.8\%$-$1.4\%$ \textit{nAUC}) when compared with those configurations without reverse reconstruction. Such results justify our claim that reverse reconstruction enables the model to learn more discriminative gait features for person Re-ID. \textbf{(c)} Introducing attention mechanism (BAS) improves the model by $6.6\%$ \textit{Rank-1} accuracy and $3.1\%$ \textit{nAUC}, while LAS further improves BAS's performance by $4.4\%$  \textit{Rank-1} and $1.2\%$ \textit{nAUC}. Besides, it is noted that directly using locality mask (MBAS) degrades the performance. This substantiates our claim that learning locality enables better gait representation learning. \textbf{(d)} AGEs provide more effective gait representations: Using AGEs can consistently improve the Re-ID performance by up to 1.8\% \textit{Rank-1} accuracy and 0.7\% \textit{nAUC}, regardless of the type of used alignment scores.
Other datasets report similar results.

\textbf{Gait Encoding Model Transfer.}  We discover that the gait encoding model learned on one dataset can be readily transferred to other datasets. For example, we use the model pre-trained on KGBD to directly encode skeletons from BIWI and IAS-Lab, and compare the Re-ID performance with the model trained on BIWI and IAS-Lab themselves (denoted by ``Self''): As shown by Table \ref{transfer}, two cases achieve fairly comparable performance, while the transferred model even outperforms the original model on IAS-A (LAS) and IAS-B (BAS). Such transferability demonstrates that our approach can learn transferable high-level semantics of 3D skeleton data, which enables the pre-trained model to capture discriminative gait features from unseen skeletons of a new dataset.

\begin{table}[t]
    \centering
    \scalebox{0.86}{
\begin{tabular}{c|cc|cc|cc}
\specialrule{0.1em}{0.45pt}{0.45pt}
\textbf{}                             & \multicolumn{2}{c|}{\textbf{BIWI}} & \multicolumn{2}{c|}{\textbf{IAS-A}} & \multicolumn{2}{c}{\textbf{IAS-B}} \\ \specialrule{0.1em}{0.45pt}{0.45pt}
\multicolumn{1}{l|}{\textbf{Config.}} & \textbf{Self}   & \textbf{Transfer} & \textbf{Self}   & \textbf{Transfer}  & \textbf{Self}   & \textbf{Transfer} \\ \specialrule{0.1em}{0.45pt}{0.45pt}
\textbf{BAS}                          & \textbf{55.1}  & 53.5              & \textbf{54.7}  & 54.2               & 56.3           & \textbf{57.1}     \\
\textbf{LAS}                          & \textbf{59.1}  & 58.4              & 56.1           & \textbf{56.3}      & \textbf{58.2}  & 57.4              \\ \specialrule{0.1em}{0.2pt}{0.2pt}
\end{tabular}
}
   \caption{\textit{Rank-1} accuracy comparison between the original model (``Self'') and the transferred model (``Transfer''). Results of different datasets and alignment score types (BAS or LAS) are reported.}
 \label{transfer}
\end{table}

\section{Conclusion}
\label{conclusions}
In this paper, we propose a generic self-supervised approach to learn effective gait representations for person Re-ID. We introduce self-supervision by learning reverse skeleton reconstruction, which enables our model to learn high-level semantics and discriminative gait features from unlabeled skeleton data. To facilitate skeleton reconstruction and gait representation learning, a novel locality-aware attention mechanism is proposed to incorporate the locality into gait encoding process. 
We construct AGEs as final gait representations to perform person Re-ID. Our approach evidently outperforms existing skeleton-based Re-ID methods, and its performance is comparable or superior to depth-based/multi-modal methods.

\section*{Acknowledgements}
This work was supported in part by the National Key Research and Development Program of China (Grant No. 2019YFA0706200, No. 2018YFB1003203), in part by the National Natural Science Foundation of China (Grant No. 61632014, No. 61627808, No. 61210010, No. 61773392, No. 61672528), in part by the National Basic Research Program of China (973 Program, Grant No. 2014CB744600), in part by the Program of Beijing Municipal Science \& Technology Commission (Grant No. Z171100000117005), and in part by the Program for Guangdong Introducing Innovative and Enterpreneurial Teams 2017ZT07X183, Fundamental Research Funds for the Central Universities D2191240. Xiping Hu, Jun Cheng and Bin Hu are the corresponding authors of this paper.

\bibliographystyle{named}
\bibliography{ijcai20}

\end{document}